\documentclass[letterpaper,10pt,conference]{ieeeconf}
\IEEEoverridecommandlockouts
\usepackage{graphicx}
\usepackage{amsmath,amssymb}
\usepackage{booktabs}
\usepackage{multirow}
\usepackage{url}
\usepackage{balance}
\usepackage{placeins}
\usepackage{hyperref}
\title{\LARGE \bf
Reliability-Aware Future Conditioning for Temporally Robust Robot Manipulation
}

\author{%
Mohammad Khoshnazar$^{1,*}$,
Mohammad Dehghani Tezerjani$^{2}$,
Zhiyuan Gao$^{1}$,
Deyuan Qu$^{3}$,\\
Max Gandyra$^{1}$,
Yanxiang Zhan$^{1}$,
Mehreen Naeem$^{1}$,
Andrew Melnik$^{1}$,\\
Jeroen Sch\"afer$^{1}$,
Qing Yang$^{2}$,
Michael Beetz$^{1}$%
\thanks{$^{1}$University of Bremen, Bremen, Germany.}%
\thanks{$^{2}$University of North Texas, Denton, TX, USA.}%
\thanks{$^{3}$Toyota Motor North America,
Mountain View, CA, USA.}%
\thanks{$^{*}$Corresponding author.
{\tt\small khoshnazar@uni-bremen.de}}%
}

\usepackage{xcolor}
\usepackage{colortbl}

\usepackage{xcolor}
\definecolor{planblue}{RGB}{0,112,192}
\definecolor{plangreen}{RGB}{0,176,80}

\begin{document}
\maketitle
\thispagestyle{empty}
\pagestyle{empty}

\begin{abstract}
A generated video of a task the robot is about to perform is useful guidance
only if it depicts the phase the robot is actually in. We show that temporal misalignment can turn
a task-consistent generated future into actively harmful guidance. 
On CALVIN, a five-frame early shift nearly erases the benefit of generated
futures, reducing success from 81.3\% to 54.8\% against 54.0\% without
futures; imposed timing shifts reduce it even further to 34.2\%, 19.8 points below the
future-free policy.
We introduce \emph{Reliability-Aware Future Conditioning} (RAFC), which
treats this as a control problem rather than a generation problem. At every
step, RAFC estimates how far to trust the received clip and which nearby
temporal hypothesis to prefer, falling back toward a static branch when neither
fits, and it learns both from task reward alone without shift labels or
alignment supervision. RAFC sits on top of \emph{Future-Experience Conditioning}
(FEC), which builds the clip once from task grounding, a robot-free digital-twin
rollout, and mask-free video diffusion. Under deliberately off-grid phase shifts and rate mismatch,
RAFC substantially improves success under temporal mismatch. Candidate ensembling accounts for most of the recovery near alignment, while learned reliability adds a further 7.0 percentage points over uniform averaging of the identical candidate bank under off-grid shifts.
 The gain holds on the evaluated task sets and
survives on a Franka under natural timing mismatch nobody imposed, where
aggregate success rises from 26.7\% to 56.7\%. All resources will be made publicly available. \href{https://future-condition.github.io/}{future-condition.github.io}.
\end{abstract}

\section{Introduction}

Future-conditioned manipulation gives a robot policy a short video of how the
task should unfold, in addition to its current observation. Such a clip exposes
what the current frame may not reveal, including the target state, motion,
upcoming contacts, and how the scene will evolve. Recent methods use generated
futures as visual subgoals~\cite{black2023susie}, exploration
targets~\cite{khoshnazar2026enact,luo2025grounding}, or replanning
feedback~\cite{bu2024closedloop}.

The key challenge is that the future signal seen during training is taken from
demonstrations and is aligned with the demonstrated motion by construction,
whereas at test time it is generated and may be temporally misaligned. At deployment, the future is generated once at task initialization at a fixed frame rate, whereas the robot's progress depends on its own dynamics and is unknown when the clip is produced. A video can therefore show the right drawer, the right direction, and the right final state, yet run several frames ahead of or behind the robot. Unlike a wrong object or goal, this error leaves the content plausible and changes only when it applies. 
It can make a generated future worse than none. 

\begin{figure}[t]
  \centering
  \includegraphics[
    width=\columnwidth,
    height=0.28\textheight,
    keepaspectratio
  ]{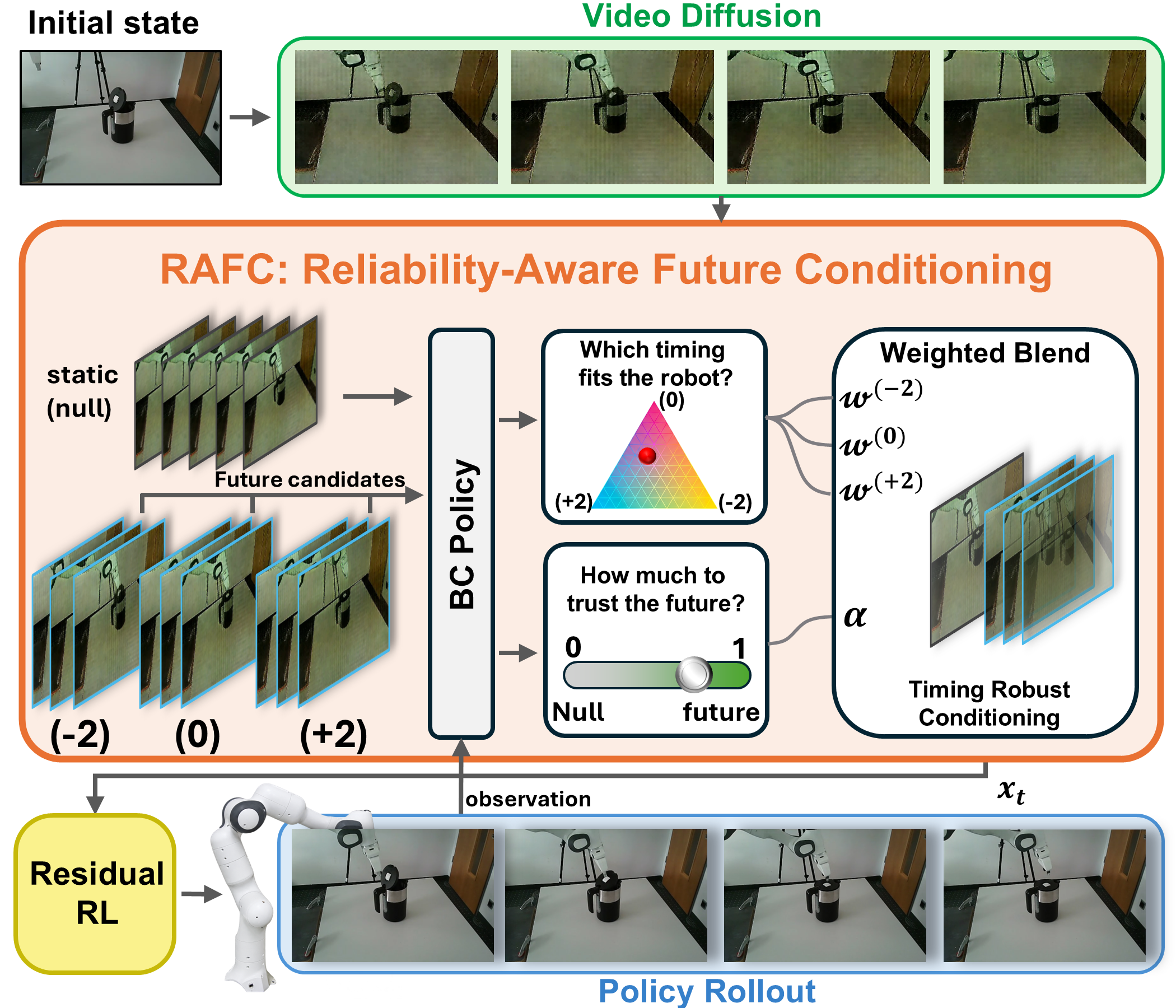}
  \caption{RAFC blends a static fallback and three temporal
  hypotheses using learned trust and phase weights, then applies
  residual control.}
  \label{fig:rafc}
\end{figure}

Solving this mismatch is difficult for three reasons. First, the robot's
phase relative to the generated video is not observed at deployment, so no
alignment label is available. Second, regenerating the future at every step is
impractical, since synthesizing one 16-frame clip takes 40--45~s on an NVIDIA
A40. Third, the mismatch varies in size and can exceed any fixed correction
range, so the controller must also recognize when the clip should not be
trusted. Existing
methods do not address these requirements. Methods that condition on generated
goals or videos target content errors and assume that the conditioning signal
is temporally compatible with execution~\cite{luo2025grounding,black2023susie}.
CLOVER~\cite{bu2024closedloop} uses embedding-space distances
to reject unreachable plans and advance between achieved subgoals, while ORCA~\cite{huey2025orca} addresses temporal misalignment in
a single demonstration used for reward shaping rather than in a future that
conditions the policy at every step.

We introduce a framework that combines \emph{Future-Experience Conditioning}
(FEC) with \emph{Reliability-Aware Future Conditioning} (RAFC). FEC constructs
the future once through language grounding, a robot-free digital-twin rollout,
and mask-free video diffusion, then encodes it for closed-loop BC and RL. RAFC
uses the frozen BC policy as a compatibility probe by evaluating a static
fallback and several time-shifted versions of the future. A learned gate
determines how strongly to trust the dynamic future and which temporal
candidate to prefer, then blends their features, latents, and base actions
before residual RL. The gate and residual actor learn jointly from task reward
without shift labels or alignment supervision.

The contributions follow in the order of the gaps above.
\begin{itemize}
\item \textbf{Temporal-failure characterization.} Controlled phase shifts,
distinct-frame source windows, and temporal-rate warps show that even
task-consistent generated futures can become actively harmful when temporally
misaligned, with severe degradation under both early and late phase errors.

\item \textbf{Reliability-aware future conditioning.} RAFC learns at every step
how strongly to rely on dynamic future information and which nearby temporal
hypothesis to prefer. Its gate and residual actor are trained jointly from task
reward without shift labels or alignment supervision.

\item \textbf{Controlled and physical evidence.} Ablations show that learned
weighting adds to shift augmentation, candidate ensembling, and a fixed learned
mixture. Gate diagnostics, an online timing intervention, CALVIN learning
curves, CALVIN and RoboCasa benchmark comparisons, and a three-task Franka
study provide controlled and physical evidence.
\end{itemize}

Section~\ref{sec:results} tests three hypotheses. \textbf{H1:} Temporal
misalignment reduces the value of task-consistent futures, and RAFC recovers
most of the resulting loss. \textbf{H2:} learned reliability provides additional recovery beyond
training-time augmentation, uniform candidate ensembling, and a fixed learned
mixture, without requiring exact candidate matching.
\textbf{H3:} RAFC improves over plain FEC on hardware under natural, unmeasured
timing variation.


\section{Related Work}

\textbf{Generated futures and temporal reliability.}
Visual anticipation predicts observations for planning~\cite{finn2017foresight},
while world models learn compact dynamics for imagined
control~\cite{ha2018world,hafner2019planet,hafner2020dreamer}. Closer to our
setting, UniPi~\cite{du2023unipi} and AVDC~\cite{ko2024avdc} generate video
plans and recover actions from them, SuSIE~\cite{black2023susie} uses
image-editing subgoals, V2A~\cite{luo2025grounding} grounds generated videos
through exploration, and FEC~\cite{khoshnazar2026enact} uses generated futures
as horizon-aware exploration hypotheses. These methods do not explicitly
estimate whether a fixed generated future remains temporally compatible with
ongoing execution. CLOVER~\cite{bu2024closedloop} replans after embedding-space
error accumulates, while ORCA~\cite{huey2025orca} aligns a single demonstration
for reward shaping. RAFC instead estimates temporal trust before each action
for a future that continuously conditions the policy.

\textbf{Temporal alignment and progress.}
Temporal Cycle Consistency (TCC) learns frame correspondences across
videos~\cite{dwibedi2019tcc}, while XIRL and VIP convert temporally structured
visual embeddings into progress-based rewards~\cite{zakka2022xirl,ma2023vip}.
ORCA~\cite{huey2025orca} aligns a learner rollout with a single misaligned
demonstration for reward shaping, whereas CLOVER~\cite{bu2024closedloop}
uses embedding-space distances for replanning and subgoal advancement. These methods use temporal
structure to define rewards or correct mismatch after it affects execution.
RAFC instead resolves timing inside the controller before each action, without
alignment supervision or an explicit progress reward, and reduces reliance on
the dynamic future when no nearby phase fits.

\textbf{Generative video, language grounding, and digital twins.}
Our generator builds on diffusion, video modeling, and
inpainting~\cite{ho2020ddpm,ho2022video,lugmayr2022repaint}, specifically
CogVideoX~\cite{yang2024cogvideox} and
VideoPainter~\cite{bian2025videopainter}. It is also related to systems using
language reasoning or imagined goals for rearrangement and affordance
grounding~\cite{kapelyukh2024dream2real,ahn2022saycan}. The generator is an
enabling component, while our study focuses on how a controller should consume
an imperfectly timed future.

\textbf{Imitation learning and RL fine-tuning.}
Our controller combines a frozen BC policy with TD3-style residual
RL~\cite{mandlekar2022robomimic, tezerjani2024rlslam, fujimoto2018td3}. Residual RL learns corrective
actions on top of an existing controller~\cite{johannink2019residual}, and
ResiP applies this idea to frozen BC policies for precise
manipulation~\cite{ankile2024resip}. RAFC retains this residual structure but
also learns how the frozen BC policy should consume its future input. The
training is related to demonstration-initialized and offline-to-online
RL~\cite{nair2020awac,lee2022offlineonline}, and we additionally evaluate
future-conditioned Streaming Flow Policy~\cite{jiang2025sfp}.

\section{Reliability-Aware Future Conditioning}
At control step $t$, the robot receives observation history $o_{t-K:t}$ and
task command $c$, and outputs a continuous 7-D action, a 6-DoF end-effector
pose delta and a gripper command. RAFC sits between a frozen
future-conditioned BC policy and residual RL, as illustrated in
Fig.~\ref{fig:rafc}. From the received future, it
forms a static fallback and three nearby temporal hypotheses, evaluates all
four with the same BC policy, and learns both how much dynamic future
information to trust and which temporal hypothesis to prefer. The resulting
weights blend the BC representations and base action before residual control.

\subsection{FEC Base Interface}
Let $v^{\mathrm{fut}}_{t,1:T_f}$ denote the future clip available at step $t$.
The BC backbone and transformer process the observation history, task command,
action-query tokens, and future tokens:
\begin{equation}
 (f_t,z^{\mathrm{fut}}_{t,1:T_f})
 =H_{\phi}(o_{t-K:t},v^{\mathrm{fut}}_{t,1:T_f},c),
 \label{eq:tokens}
\end{equation}
where $f_t$ is the action-token policy feature and
$z^{\mathrm{fut}}_{t,1:T_f}$ are the future-token representations. We preserve
coarse temporal structure by pooling the future tokens into $B=4$ bins and
projecting their concatenation:
\begin{equation}
 u_{t,1:B}=\mathrm{Pool}_{B}(z^{\mathrm{fut}}_{t,1:T_f}),\qquad
 g_t=P([u_{t,1};\ldots;u_{t,B}]).
 \label{eq:futurelatent}
\end{equation}
The BC head predicts
$a_t^{\mathrm{BC}}=\pi^{\mathrm{BC}}_\phi(f_t,g_t)$ and is trained from
demonstration tuples
$(o_{t-K:t},v^{\mathrm{fut}}_{t,1:T_f},c,a_t^\star)$ with
\begin{equation}
 \mathcal L_{\mathrm{BC}}(\phi)=\mathbb E\!\left[
 \left\|\pi^{\mathrm{BC}}_\phi(f_t,g_t)-a_t^\star\right\|_2^2\right].
 \label{eq:bcobjective}
\end{equation}
BC uses demonstration futures during training and generated futures at
inference. For each candidate future, its frozen interface exposes the policy
feature $f_t$, pooled future latent $g_t$, and BC base action
$a_t^{\mathrm{BC}}$.

\subsection{Reliability-Aware Future Conditioning}
RAFC uses an ordered bank with one static null clip and three dynamic
candidates obtained by local temporal offsets $\mathcal S=\{-2,0,+2\}$ of the
received future, where $s=0$ reproduces the received clip unchanged. The null branch
repeats the first frame,
\begin{equation}
 v^{\mathrm{null}}_{t,i}=v^{\mathrm{fut}}_{t,1},\qquad i=1,\ldots,T_f,
\end{equation}
retaining scene and task appearance while removing temporal evolution. It is
therefore a static fallback, not NoFuture, which removes future guidance
entirely. The dynamic candidates apply these offsets to the received future,
\begin{equation}
 v^{(s)}_{t,i}=v^{\mathrm{fut}}_{t,\mathrm{clip}(i+s,1,T_f)},
 \qquad s\in\mathcal S,
 \label{eq:localshift}
\end{equation}
with boundary frames repeated only when a local index falls outside the clip.

All four candidates are present at every RAFC training transition; their local
offsets are fixed and never provided as labels. Larger global perturbations are
used only for evaluation and are never supplied to RAFC as inputs or labels.
The frozen BC policy produces
$(f_t^{\mathrm{null}},g_t^{\mathrm{null}},a_t^{\mathrm{BC,null}})$ for the
static branch and
$(f_t^{(s)},g_t^{(s)},a_t^{\mathrm{BC},(s)})$ for $s\in\mathcal S$.

The gate receives only the fixed-order BC policy features,
\begin{equation}
  h_t=\operatorname{concat}\!\left(
 f_t^{\mathrm{null}},f_t^{(-2)},f_t^{(0)},f_t^{(+2)}
 \right)\in\mathbb R^{4d_f},
\end{equation}
where $d_f=\dim(f_t)$. It outputs one trust logit and three phase logits,
\[
(\ell_t^\alpha,\ell_t^{-2},\ell_t^{0},\ell_t^{+2})=G_\psi(h_t),
\]
which define
\begin{align}
\alpha_t &= \sigma(\ell_t^\alpha),\\
w_{t,s} &=
\frac{\exp(\ell_t^s)}
{\sum_{s'\in\mathcal S}\exp(\ell_t^{s'})},
\qquad s\in\mathcal S.
\end{align}
Here $\alpha_t$ assigns mass between the static fallback and dynamic future,
while $w_{t,s}$ distributes the dynamic mass among candidate phases.

The same coefficients blend all three BC outputs:
\begin{align}
f_t^{\mathrm{gate}}&=(1-\alpha_t)f_t^{\mathrm{null}}
+\alpha_t\!\sum_{s\in\mathcal S}w_{t,s}f_t^{(s)},\\
g_t^{\mathrm{gate}}&=(1-\alpha_t)g_t^{\mathrm{null}}
+\alpha_t\!\sum_{s\in\mathcal S}w_{t,s}g_t^{(s)},\label{eq:gate}\\
a_t^{\mathrm{BC,gate}}&=(1-\alpha_t)a_t^{\mathrm{BC,null}}
+\alpha_t\!\sum_{s\in\mathcal S}w_{t,s}a_t^{\mathrm{BC},(s)}.
\end{align}
Thus $\alpha_t$ answers whether dynamic future information should be trusted,
whereas $w_{t,s}$ selects the most compatible nearby temporal hypothesis.


The gate receives neither the applied global shift nor a temporal-alignment
target. Candidate identity is represented only by fixed position in $h_t$.
Because all branches share the task and scene, compatibility with the current
observation must be inferred from frozen-BC features. The gate is learned
jointly with the residual actor from the actor objective of
Eq.~\eqref{eq:actorloss} alone.

\subsection{RAFC Training with Residual RL}
Training has three stages. We first train the video diffusion model on paired
twin/robot clips with first-frame anchoring, then train the future-conditioned
BC policy with Eq.~\eqref{eq:bcobjective} on demonstration futures. RAFC is
absent from both stages. We then freeze BC and optimize only the RAFC gate,
bounded residual actor, and twin critics during off-policy RL fine-tuning.

The residual actor and executed action are
\begin{equation}
\Delta a_t=\pi_\theta(f_t^{\mathrm{gate}},g_t^{\mathrm{gate}},
a_t^{\mathrm{BC,gate}}),\quad
a_t=a_t^{\mathrm{BC,gate}}+\Delta a_t.
\label{eq:residual}
\end{equation}
The critics use
$x_t=(f_t^{\mathrm{gate}},g_t^{\mathrm{gate}},a_t^{\mathrm{BC,gate}})$.
On delayed policy updates, the actor objective trains both the residual actor
and the gate:
\begin{equation}
 \mathcal L_\pi(\theta,\psi)=-\mathbb E\!\left[
 Q_{\omega_1}\!\left(x_t,
 \pi_\theta(f_t^{\mathrm{gate}},g_t^{\mathrm{gate}},
 a_t^{\mathrm{BC,gate}})\right)\right].
 \label{eq:actorloss}
\end{equation}
The RL reward contains only a step cost, residual-action penalty, and sparse
terminal outcomes:
\begin{align}
 \rho_t={}&r_{\mathrm{step}}
 -\lambda_\Delta\|\Delta a_t^{\mathrm{arm}}\|_2^2
 +r_{\mathrm{terminal}},\\
 r_t={}&\mathrm{clip}(\rho_t,-50,150).
 \label{eq:reward}
\end{align}
We set $r_{\mathrm{step}}=-0.005$ and $\lambda_\Delta=0.02$, with terminal
rewards $+120/-25/-20$ for success, timeout, and broken steps. No shift label,
alignment score, reference trajectory, or simulator-state distance is used.
At evaluation, the critics are discarded; only frozen BC, the RAFC gate, and
the residual actor remain.

\newcounter{rafcalgorithm}
\begin{figure}[!htbp]
\centering
\refstepcounter{rafcalgorithm}\label{alg:rafc}
\begin{minipage}{\columnwidth}
\scriptsize
\hrule
\vspace{2pt}
\textbf{Algorithm \therafcalgorithm}\quad RAFC TD3-style residual fine-tuning
\vspace{2pt}
\hrule
\vspace{2pt}
\textbf{Require:} Frozen BC policy, RAFC gate $G_\psi$, residual actor $\pi_\theta$, critics $Q_{\omega_1},Q_{\omega_2}$

\textbf{Require:} Replay buffer $\mathcal B$, shifts $\mathcal S=\{-2,0,+2\}$

\setlength{\tabcolsep}{0pt}
\renewcommand{\arraystretch}{1.05}
\begin{tabular}{@{}r@{\hspace{0.45em}}p{0.91\columnwidth}@{}}
1: & \textbf{for} each environment step $t$ \textbf{do} \\
2: & \hspace*{1em}Construct the null and $\{-2,0,+2\}$ future candidates and run each through the frozen BC policy \\
3: & \hspace*{1em}Concatenate candidate policy features into $h_t$ and compute $\alpha_t$ and $w_{t,s}$ with $G_\psi(h_t)$ \\
4: & \hspace*{1em}Blend candidate policy features, future latents, and BC actions using $\alpha_t$ and $w_{t,s}$ \\
5: & \hspace*{1em}Compute $\Delta a_t=\pi_\theta(\cdot)+\epsilon_t$ and execute $a_t=a_t^{\mathrm{BC,gate}}+\Delta a_t$ \\
6: & \hspace*{1em}Observe the next transition and store it in $\mathcal B$ \\
7: & \hspace*{1em}\textbf{for} each gradient update \textbf{do} \\
8: & \hspace*{2em}Sample a minibatch; compute target gated representations and target residual actions \\
9: & \hspace*{2em}Update twin critics with the TD3 clipped double-Q objective \\
10: & \hspace*{2em}\textbf{if} delayed policy update \textbf{then} \\
11: & \hspace*{3em}Update the residual actor and RAFC gate by minimizing $\mathcal L_\pi$ \\
12: & \hspace*{3em}Update target networks by exponential moving average \\
13: & \hspace*{2em}\textbf{end if} \\
14: & \hspace*{1em}\textbf{end for}; \textbf{end for} \\
\end{tabular}
\vspace{2pt}
\hrule
\end{minipage}
\end{figure}

\subsection{Future-Generation Testbed}
\begin{figure*}[!t]
  \centering
  \includegraphics[width=0.96\textwidth]{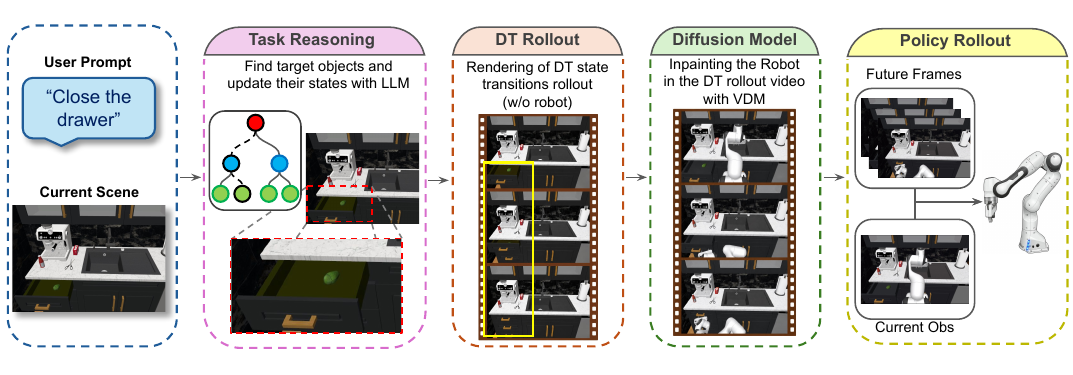}
  \caption{Future-generation testbed used by FEC and RAFC. An ontology-guided LLM specifies the target object, interaction part, desired state transition, and rollout constraints. A robot-free digital-twin rollout makes the intended scene change explicit, and mask-free video diffusion reconstructs the robot-containing future supplied to the policy.}
  \label{fig:pipeline}
\end{figure*}

Given command $c$ and scene state once at task initialization, an ontology-guided LLM (GPT-4o) returns the manipulated object, interaction part, desired state transition, and rollout constraints in a fixed JSON schema. The digital twin resolves that object to a simulator body and articulation joint. For articulated tasks, it linearly interpolates the corresponding joint coordinate between its observed and commanded limits; for planar object-motion tasks, it interpolates the object pose toward the task-defined goal region. Rendering these states produces a robot-free rollout $\tilde v_{1:T}$
(Fig.~\ref{fig:pipeline}).

Omitting the robot separates the desired scene transition from
embodiment-specific motion, so the twin need not synthesize a dynamically valid
robot trajectory while object motion stays explicit. A mask-free video
diffusion model then reconstructs a robot-containing future $v_{1:T}$ from the
initial robot image and transparent rollout, using a CogVideoX backbone with a
VideoPainter-style conditioning branch~\cite{yang2024cogvideox,bian2025videopainter}.
The future is constructed once at initialization, so grounding cost is not paid
per control step.

\section{Experiments}
\subsection{Benchmarks and Tasks}
We evaluate eight CALVIN~\cite{mees2022calvin} tasks and five
RoboCasa~\cite{nasiriany2024robocasa} tasks. We select state-changing tasks
with observable target articulation or object displacement, covering
prismatic, revolute, toggle, and planar interactions while admitting a
deterministic digital-twin specification. Simulation permits future source
and temporal alignment to change while task state remains fixed.
 The two benchmarks play different roles. CALVIN carries the temporal-robustness analysis, since its eight tasks support the controlled phase, window, and rate interventions of Sec. IV-B; RoboCasa tests whether the system-level gain transfers to a different scene distribution and task suite under unperturbed futures.

\subsection{Temporal-Misalignment Protocol}
To evaluate temporal robustness without changing task identity or
initialization, we perturb the received future only at evaluation. For a
$T$-frame clip, a global phase shift $s$ is applied before future-token pooling:
\begin{equation}
 v^{(s)}_{i}=v_{\mathrm{clip}(i+s,1,T)},
 \label{eq:shift}
\end{equation}
where out-of-range indices repeat the boundary frame. We evaluate deliberately off-grid shifts
$s\in\{-5,-3,-1,0,+1,+3,+5\}$, so no nonzero evaluation perturbation
coincides with a local RAFC candidate in $\mathcal S=\{-2,0,+2\}$. This
intervention preserves task identity and motion direction while changing the
interaction phase represented by the future. For RAFC, the local candidate bank
of Eq.~\eqref{eq:localshift} is constructed \emph{after} this global
perturbation; the applied global shift is never given to the gate.

Because large shifts in Eq.~\eqref{eq:shift} can reduce temporal diversity
through boundary repetition, we also use an independently generated
distinct-frame protocol. For each evaluation instance, we produce a separate
28-frame source rollout $\bar v_{1:28}$ and extract a contiguous 16-frame
window,
\begin{equation}
 v^{\mathrm{win}(s)}_i=\bar v_{i+6+s},\qquad i=1,\ldots,16,
 \label{eq:windowshift}
\end{equation}
using the same off-grid evaluation set. At $s=\pm5$, every window contains
16 distinct frames, spanning frames 2--17 through 12--27. The windowed source,
including its $s=0$ window, is independent of the original 16-frame clip used
in the imposed-shift protocol; no result is reused between the two protocols.
This isolates phase mismatch from clipping, padding, and repeated boundary
frames in the imposed global perturbation. We additionally retain $s=\pm6$
only as an extreme stress test for the failure magnitude; these two conditions
are excluded from the off-grid averages and candidate-matching analysis.

We separately test rate mismatch by time-warping the same source rollout with
$j_i=\mathrm{round}(1+\rho(i-1))$, $\rho\in\{0.75,1.25\}$, using enough source
frames to avoid padding. Here the phase error grows with horizon. These global
phase shifts, distinct-frame windows, and rate warps are evaluation-only and
unknown to RAFC.

For the online intervention, episodes start at $s=0$ and switch without warning
at step 40 to $s\in\{-5,-3,+3,+5\}$, applied to the original clip and held to
termination; all network weights remain frozen and the offset is hidden.

\subsection{Conditions, Controllers, and Training}
\textbf{NoFuture} removes guidance, \textbf{GTFuture} provides an oracle
demonstration clip, and \textbf{GenFuture} uses the generated clip. We compare
BC-only, BC+RL, and future-conditioned SFP~\cite{jiang2025sfp}. \textbf{ShiftAug} samples a fixed per-episode offset from $\{-2,0,+2\}$ during
RL, using the same frozen BC policy and budget, no offset label, and a single
future without RAFC or a null branch. Temporal-rate warps are evaluation-only.

The online-change study compares full RAFC with two matched-budget controls:
\textbf{ConstantGate}, which learns one observation-independent $\alpha$ and
three temporal weights jointly with its own residual actor, and \textbf{Uniform
averaging} over the same four branches. Initial states and disturbances are
matched across methods and seeds 42, 43, and 44.

We additionally evaluate against \textbf{Diffusion Policy}~\cite{chi2023diffusion}, \textbf{SuSIE}~\cite{black2023susie}, \textbf{CLOVER}~\cite{bu2024closedloop}, and \textbf{V2A}~\cite{luo2025grounding}, using the official implementations released by the respective authors. In the standard baseline comparisons of Tables~\ref{tab:sota_calvin} and~\ref{tab:sota_robocasa}, all four methods, including CLOVER, are evaluated under unshifted conditions. Table~\ref{tab:shift} is a separate temporal-robustness experiment in which, among these four baselines, only CLOVER is additionally evaluated under the same imposed shifts as our method. We do not report shifted results for Diffusion Policy, SuSIE, or V2A because Diffusion Policy has no future-conditioning signal, SuSIE uses a single image subgoal, and V2A couples video generation with action grounding inside its own pipeline, making the same frame-level intervention inapplicable.

Each future contains $T=16$ frames and is compressed into $B=4$ temporal bins.
BC and BC+RL use ResNet-18 visual encoding followed by a transformer. Standard
RL fine-tuning uses batch size 256, discount $\gamma=0.99$, target update
$\tau=0.005$, policy delay 2, actor/critic learning rates $5\times10^{-5}$,
target policy noise 0.05, noise clip 0.08, and 140k environment steps. We
train three independent seeds (42, 43, 44) for each controller and condition
and evaluate 500 simulation episodes per task and seed; episodes last at most
200 steps. We report means and standard deviations across the three
seed-level success rates and use task-balanced benchmark aggregates.

\subsection{Future Construction and Reproducibility}
At task initialization, GPT-4o maps the command and initial scene state to the fixed grounding schema specifying the manipulated object, interaction part, desired state change, and rollout constraints. The digital twin then renders the corresponding robot-free object motion. Together with the initial robot observation, this rollout conditions the VDM to synthesize the 16-frame robot-containing future supplied to the controller. GPT-4o does not generate low-level actions and is never queried during closed-loop execution.

For VDM training, simulation provides robot-free clips directly from the renderer. For real-world data, we use ProPainter~\cite{zhou2023propainter} to remove the robot arm and gripper from recorded demonstrations and reconstruct the occluded background, yielding robot-free videos. ProPainter is used only for offline data preparation and is not part of the deployment pipeline. Training uses fixed windows and first-frame anchoring, and GenFuture never falls back to a demonstration clip. On one NVIDIA A40, a 16-frame clip with 20 denoising steps takes 40--45~s in bf16, once at initialization. Online RAFC inference takes 120~ms per control step (8.3~Hz).


\section{Results}
\label{sec:results}
\subsection{H1: Timing Mismatch Can Make Generated Futures Actively Harmful}
Table~\ref{tab:shift} establishes the failure mode and the recovery. A
task-consistent generated future is worth 15.8 points at $s=0$, but the same
clip drops GenFuture to 34.2 at $s=-5$ and 41.6 at $s=+5$, below the
future-free NoFuture reference of 54.0. An additional extreme stress test at
$s=-6/+6$ gives 25.7/42.6, so temporal misalignment can cost up to 28.3 points
relative to using no future at all. 

GTFuture is also sensitive to timing: its shifted average falls
from 89.1\% at zero shift to 75.2\%, a loss of 13.9 percentage
points, and its largest loss is 27.0 points at $s=-5$.
However, GTFuture remains above NoFuture at every evaluated
off-grid offset, whereas GenFuture falls below NoFuture at
both $s=\pm5$ and $s=\pm3$.

RAFC raises the off-grid shifted average from 51.8\% to 73.7\% and improves
every shifted condition without seeing the applied offset. It exceeds NoFuture
at $s=-5,-3,-1,0,+1,+3,+5$; at $\pm5$, the $\pm2$ candidate range leaves a residual
three-frame mismatch, requiring reduced dynamic-future trust. Consistently,
RAFC's 61.7\%/65.2\% at global $\pm5$ lies near or above GenFuture's
52.9\%/51.9\% at $\pm3$.


Independent distinct-frame source windows reproduce the pattern (69.7\% to
75.6\%), so the gain is not an artifact of boundary repetition or reuse of the
original $s=0$ clip. Under rate warps, GenFuture reaches 57.1\%/60.1\% at
$0.75\times/1.25\times$ while RAFC reaches 70.9\%/75.4\%
(58.6\%/73.2\% average, Table~\ref{tab:ratewarp}), against a shift-invariant
NoFuture of 54.0\%.

\begin{table*}[!t]
\centering
\caption{Off-grid CALVIN temporal perturbation, mean $\pm$ s.d.
success (\%) over three seeds. No nonzero shift matches the RAFC
bank $\{-2,0,+2\}$; windowed rows use independent sources.
Avg.\ shifted excludes zero shift; bold highlights the RAFC rows.}
\label{tab:shift}
\scriptsize
\setlength{\tabcolsep}{3pt}
\renewcommand{\arraystretch}{1.15}
\begin{tabular*}{\textwidth}{
  @{\extracolsep{\fill}}l*{8}{c}@{}
}
\toprule
Future source / Method
& $-5$ & $-3$ & $-1$ & $0$ & $+1$ & $+3$ & $+5$
& \shortstack{Avg.\\shifted}\\
\midrule
CLOVER~\cite{bu2024closedloop}
& $42.8\pm5.0$
& $66.1\pm4.2$
& $74.3\pm2.9$
& $71.9\pm2.4$
& $74.0\pm3.1$
& $66.7\pm4.4$
& $43.9\pm5.2$
& $61.3$\\
\midrule
NoFuture
& \textcolor{gray}{$54.0\pm3.2$}
& \textcolor{gray}{$54.0\pm3.2$}
& \textcolor{gray}{$54.0\pm3.2$}
& \textcolor{gray}{$54.0\pm3.2$}
& \textcolor{gray}{$54.0\pm3.2$}
& \textcolor{gray}{$54.0\pm3.2$}
& \textcolor{gray}{$54.0\pm3.2$}
& \textcolor{gray}{$54.0$}\\
GenFuture
& $34.2\pm4.7$
& $52.9\pm3.5$
& $64.6\pm2.6$
& $69.8\pm2.1$
& $65.3\pm2.9$
& $51.9\pm3.6$
& $41.6\pm5.1$
& $51.8$\\
\textbf{GenFuture+RAFC}
& $\mathbf{61.7\pm4.9}$
& $\mathbf{76.8\pm3.7}$
& $\mathbf{81.2\pm2.4}$
& $\mathbf{82.3\pm1.6}$
& $\mathbf{81.5\pm2.6}$
& $\mathbf{76.0\pm4.1}$
& $\mathbf{65.2\pm5.0}$
& $\mathbf{73.7}$\\
\midrule
GenFuture (windowed)
& $54.8\pm4.5$
& $72.9\pm3.9$
& $79.4\pm2.8$
& $81.3\pm2.6$
& $79.8\pm3.1$
& $72.4\pm3.7$
& $58.9\pm5.2$
& $69.7$\\
\textbf{GenFuture+RAFC (windowed)}
& $\mathbf{64.9\pm4.6}$
& $\mathbf{77.6\pm3.2}$
& $\mathbf{82.1\pm2.7}$
& $\mathbf{83.7\pm2.0}$
& $\mathbf{82.0\pm3.0}$
& $\mathbf{78.0\pm3.9}$
& $\mathbf{69.1\pm4.7}$
& $\mathbf{75.6}$\\
\midrule
GTFuture (oracle)
& $62.1\pm3.9$
& $78.4\pm2.8$
& $85.3\pm1.9$
& $89.1\pm1.6$
& $84.3\pm2.1$
& $76.8\pm3.4$
& $64.1\pm4.2$
& $75.2$\\
\bottomrule
\end{tabular*}
\end{table*}

\subsection{Unshifted Futures Establish the Base Case}
Table~\ref{tab:controllers} shows that future conditioning helps BC-only and
BC+RL on CALVIN. Under BC+RL, GenFuture improves over NoFuture by 15.8 points
(69.8 vs.\ 54.0). GTFuture reaches 89.1\%, exceeding GenFuture by 19.3 points;
this gap reflects differences in future content and temporal compatibility.
RL fine-tuning achieves the highest success under every future source.

\begin{table}[!htbp]
\caption{CALVIN task-balanced success (\%), mean $\pm$ s.d. over three seeds.
GTFuture is oracle; bold marks the highest mean.}
\label{tab:controllers}
\centering
\scriptsize
\setlength{\tabcolsep}{4.0pt}
\begin{tabular}{lcccc}
\toprule
Controller & GT & Gen & NoF & $\Delta$ Gen--NoF \\
\midrule
SFP
& $24.3\!\pm\!4.7$
& $17.8\!\pm\!2.9$
& $6.1\!\pm\!4.1$
& $+11.7$ \\
BC-only
& $61.1\!\pm\!0.8$
& $53.6\!\pm\!1.9$
& $39.8\!\pm\!4.3$
& $+13.8$ \\
BC+RL
& $\mathbf{89.1\!\pm\!1.6}$
& $\mathbf{69.8\!\pm\!2.1}$
& $\mathbf{54.0\!\pm\!3.2}$
& $\mathbf{+15.8}$ \\
\bottomrule
\end{tabular}
\end{table}

\subsection{Comparison with Prior Methods}
Tables~\ref{tab:sota_calvin} and~\ref{tab:sota_robocasa} report the
CALVIN and RoboCasa comparisons. The full FEC+RAFC system achieves the highest
non-oracle average on both evaluated task sets, reaching 82.3\% on CALVIN and
76.6\% on RoboCasa, while CLOVER remains best on close-drawer and slider in
CALVIN and TurnOnMicrowave in RoboCasa.

\begin{table*}[!t]
\centering
\caption{Comparison on the evaluated CALVIN task set: success rate (\%),
reported as mean $\pm$ standard deviation. Bold indicates the best
non-oracle method for each task.
$\dagger$V2A uses self-exploration rather than
demonstration-initialized BC+RL.}
\label{tab:sota_calvin}
\scriptsize
\setlength{\tabcolsep}{3pt}
\renewcommand{\arraystretch}{1.25}
\begin{tabular*}{\textwidth}{
@{\extracolsep{\fill}}l*{9}{c}@{}
}
\toprule
Method
& \shortstack{Open\\drawer}
& \shortstack{Close\\drawer}
& \shortstack{Bulb\\on}
& \shortstack{Bulb\\off}
& \shortstack{LED\\on}
& \shortstack{LED\\off}
& Push
& Slider
& Avg. \\
\midrule

Diffusion Policy~\cite{chi2023diffusion}
& $43.1\!\pm\!1.4$ & $44.4\!\pm\!2.1$
& $38.9\!\pm\!0.9$ & $31.2\!\pm\!1.2$
& $40.8\!\pm\!2.1$ & $32.1\!\pm\!0.9$
& $17.1\!\pm\!1.2$ & $25.0\!\pm\!1.6$
& $34.1\!\pm\!1.2$ \\

SuSIE~\cite{black2023susie}
& $52.1\!\pm\!3.7$ & $57.7\!\pm\!2.2$
& $72.5\!\pm\!1.4$ & $58.4\!\pm\!2.6$
& $71.4\!\pm\!5.8$ & $76.8\!\pm\!2.4$
& $3.6\!\pm\!1.3$ & $5.1\!\pm\!0.4$
& $49.7\!\pm\!1.9$ \\

V2A$^\dagger$~\cite{luo2025grounding}
& $67.6\!\pm\!1.3$ & $61.4\!\pm\!1.7$
& $69.3\!\pm\!1.6$ & $63.4\!\pm\!3.1$
& $66.4\!\pm\!2.8$ & $73.1\!\pm\!3.2$
& $19.2\!\pm\!0.3$ & $14.5\!\pm\!1.3$
& $54.4\!\pm\!1.5$ \\

CLOVER~\cite{bu2024closedloop}
& $89.6\!\pm\!2.3$ & $\mathbf{91.2\!\pm\!0.7}$
& $71.9\!\pm\!5.2$ & $75.1\!\pm\!3.3$
& $87.4\!\pm\!1.2$ & $89.9\!\pm\!3.1$
& $23.1\!\pm\!1.1$ & $\mathbf{46.9\!\pm\!2.1}$
& $71.9\!\pm\!2.4$ \\

\midrule

\shortstack[l]{\textbf{FEC+RAFC}\\\textbf{(GenFuture)}}
& $\mathbf{98.5\!\pm\!0.9}$ & $90.2\!\pm\!1.1$
& $\mathbf{94.8\!\pm\!2.2}$ & $\mathbf{95.4\!\pm\!0.9}$
& $\mathbf{92.0\!\pm\!2.1}$ & $\mathbf{93.7\!\pm\!2.1}$
& $\mathbf{51.5\!\pm\!1.2}$ & $42.6\!\pm\!2.1$
& $\mathbf{82.3\!\pm\!1.6}$ \\

\bottomrule
\end{tabular*}
\end{table*}

\begin{table*}[!t]
\caption{RoboCasa task success (\%), mean $\pm$ s.d. over three seeds.
Avg. averages task means; bold marks the best non-oracle method;
$\dagger$V2A uses self-exploration.}
\label{tab:sota_robocasa}
\centering
\scriptsize
\setlength{\tabcolsep}{3.8pt}
\begin{tabular}{lcccccc}
\toprule
Method
& CloseDrawer
& OpenSingleDoor
& OpenDrawer
& TurnOnMicrowave
& PickPlaceSinkToCounter
& Avg. \\
\midrule

Diffusion Policy~\cite{chi2023diffusion}
& $39.1\!\pm\!2.9$
& $42.1\!\pm\!1.8$
& $37.8\!\pm\!1.2$
& $54.8\!\pm\!4.1$
& $22.6\!\pm\!2.8$
& $39.3$ \\

SuSIE~\cite{black2023susie}
& $62.7\!\pm\!3.1$
& $57.4\!\pm\!3.6$
& $68.9\!\pm\!3.9$
& $71.8\!\pm\!1.9$
& $56.7\!\pm\!2.1$
& $63.5$ \\

V2A$^\dagger$~\cite{luo2025grounding}
& $68.9\!\pm\!2.6$
& $63.5\!\pm\!3.0$
& $61.8\!\pm\!2.4$
& $69.1\!\pm\!4.1$
& $50.7\!\pm\!2.8$
& $62.8$ \\

CLOVER~\cite{bu2024closedloop}
& $72.1\!\pm\!3.2$
& $69.6\!\pm\!3.2$
& $65.9\!\pm\!1.3$
& $\mathbf{84.0\!\pm\!3.2}$
& $71.3\!\pm\!1.9$
& $72.6$ \\

\midrule

FEC+RAFC, GenFuture
& $\mathbf{80.1\!\pm\!2.3}$
& $\mathbf{74.3\!\pm\!2.8}$
& $\mathbf{77.1\!\pm\!2.6}$
& $78.2\!\pm\!1.1$
& $\mathbf{73.1\!\pm\!3.1}$
& $\mathbf{76.6}$ \\

\bottomrule
\end{tabular}
\end{table*}

\subsection{H2: Learned Reliability Adds Beyond Ensembling}
\label{sec:ablations}
Table~\ref{tab:rafcfull} separates the mechanism from its ingredients.
Training-time augmentation over the same local offsets (ShiftAug) reaches 65.1,
uniform averaging over the same candidate bank reaches 66.7, learned weighting
without the null branch reaches 71.3, and full RAFC reaches 73.7. Candidate
ensembling therefore accounts for part of the effect, while learned weighting
and the static fallback add 7.0 points beyond uniform averaging on the off-grid
evaluation set.

Table~\ref{tab:gatediag} shows the two gate levels behaving independently,
dynamic-future trust decreases as mismatch grows, while candidate weighting
moves in the compensating direction even though none of the evaluated nonzero
offsets exactly matches a local candidate.

\begin{table}[!t]
\caption{RAFC gate diagnostics under off-grid temporal shifts
(mean $\pm$ s.d. over three seed-level averages).}
\label{tab:gatediag}
\centering
\scriptsize
\setlength{\tabcolsep}{2.6pt}
\begin{tabular}{lcccc}
\toprule
Condition & $\alpha$ & $w_{-2}$ & $w_0$ & $w_{+2}$\\
\midrule
Unshifted
& $.79\!\pm\!.08$ & $.18\!\pm\!.05$ & $.59\!\pm\!.10$ & $.23\!\pm\!.07$\\
Shift $-3$
& $.61\!\pm\!.11$ & $.14\!\pm\!.05$ & $.34\!\pm\!.10$ & $.52\!\pm\!.11$\\
Shift $+3$
& $.64\!\pm\!.10$ & $.54\!\pm\!.12$ & $.32\!\pm\!.09$ & $.14\!\pm\!.05$\\
Shift $-5$
& $.41\!\pm\!.12$ & $.11\!\pm\!.04$ & $.25\!\pm\!.09$ & $.64\!\pm\!.13$\\
Shift $+5$
& $.45\!\pm\!.11$ & $.62\!\pm\!.12$ & $.27\!\pm\!.08$ & $.11\!\pm\!.04$\\
\bottomrule
\end{tabular}
\end{table}

\textbf{Online timing intervention.} Table~\ref{tab:online_shift} records
$\alpha_t$, effective correction $\bar{s}_t=\sum_s w_{t,s}s$, which is a convex combination over S and therefore bounded by two frames, 
and matched
success after unannounced off-grid step-40 changes. The $\pm3$ conditions
require compensation without an exact candidate match, while $\pm5$
additionally probe trust reduction outside the local correction range.

\begin{table*}[!t]
\caption{Training-time augmentation and RAFC ablation under off-grid temporal
shifts on CALVIN. Mean $\pm$ s.d. over three seeds; bold indicates the highest
mean in each column.}
\label{tab:rafcfull}
\centering
\scriptsize
\setlength{\tabcolsep}{5.2pt}
\begin{tabular}{lcccccccc}
\toprule
Method & $-5$ & $-3$ & $-1$ & $0$ & $+1$ & $+3$ & $+5$ & Avg. shifted\\
\midrule
Forced GenFuture
& $34.2\!\pm\!4.7$
& $52.9\!\pm\!3.5$
& $64.6\!\pm\!2.6$
& $69.8\!\pm\!2.1$
& $65.3\!\pm\!2.9$
& $51.9\!\pm\!3.6$
& $41.6\!\pm\!5.1$
& $51.8$\\
GenFuture+ShiftAug
& $46.7\!\pm\!3.0$
& $67.4\!\pm\!2.5$
& $78.0\!\pm\!1.7$
& $80.5\!\pm\!2.6$
& $78.6\!\pm\!3.2$
& $67.0\!\pm\!3.0$
& $53.0\!\pm\!2.2$
& $65.1$\\
Uniform avg.
& $49.6\!\pm\!1.5$
& $68.9\!\pm\!2.8$
& $78.4\!\pm\!0.8$
& $79.4\!\pm\!1.8$
& $79.0\!\pm\!1.0$
& $68.8\!\pm\!2.1$
& $55.5\!\pm\!2.4$
& $66.7$\\
RAFC w/o null
& $57.4\!\pm\!2.1$
& $74.0\!\pm\!3.3$
& $80.3\!\pm\!1.9$
& $81.1\!\pm\!0.8$
& $80.6\!\pm\!1.3$
& $73.6\!\pm\!2.7$
& $61.8\!\pm\!2.2$
& $71.3$\\
Full RAFC
& $\mathbf{61.7\!\pm\!4.9}$
& $\mathbf{76.8\!\pm\!3.7}$
& $\mathbf{81.2\!\pm\!2.4}$
& $\mathbf{82.3\!\pm\!1.6}$
& $\mathbf{81.5\!\pm\!2.6}$
& $\mathbf{76.0\!\pm\!4.1}$
& $\mathbf{65.2\!\pm\!5.0}$
& $\mathbf{73.7}$\\
\bottomrule
\end{tabular}
\end{table*}

\begin{table}[!t]
\caption{Step-40 off-grid timing intervention on CALVIN. RAFC means are
before$\rightarrow$after; U/C/R are Uniform/ConstantGate/RAFC success (\%).}
\label{tab:online_shift}
\centering
\scriptsize
\resizebox{\columnwidth}{!}{%
\begin{tabular}{lccccc}
\toprule
Change & $\alpha$ & $\bar{s}$ & U & C & R \\
\midrule
$0\!\rightarrow\!-3$
& $.79\!\rightarrow\!.63$
& $.12\!\rightarrow\!+.92$
& $72.8\!\pm\!2.7$
& $75.1\!\pm\!2.5$
& $\mathbf{80.3\!\pm\!2.0}$ \\
$0\!\rightarrow\!+3$
& $.79\!\rightarrow\!.65$
& $.12\!\rightarrow\!-.88$
& $72.1\!\pm\!2.6$
& $74.6\!\pm\!2.9$
& $\mathbf{79.7\!\pm\!2.3}$ \\
$0\!\rightarrow\!-5$
& $.78\!\rightarrow\!.43$
& $.11\!\rightarrow\!+1.19$
& $56.2\!\pm\!4.0$
& $61.0\!\pm\!3.8$
& $\mathbf{69.4\!\pm\!3.7}$ \\
$0\!\rightarrow\!+5$
& $.80\!\rightarrow\!.46$
& $.13\!\rightarrow\!-1.13$
& $61.5\!\pm\!3.7$
& $64.8\!\pm\!4.1$
& $\mathbf{72.0\!\pm\!3.5}$ \\
\midrule
Avg. success & -- & -- & $65.7$ & $68.9$ & $\mathbf{75.4}$ \\
\bottomrule
\end{tabular}}
\end{table}

Full RAFC averages 75.4\% success versus 68.9\% for ConstantGate and 65.7\%
for uniform averaging. Its step-40 response compensates through temporal
reweighting under $\pm3$, while $\pm5$ additionally produces a marked reduction
in $\alpha$, shifting mass toward the static fallback.

\begin{table}[!htbp]
\caption{Robustness to temporal-rate mismatch on CALVIN (mean $\pm$ s.d. over three seeds).}
\label{tab:ratewarp}
\centering
\scriptsize
\setlength{\tabcolsep}{3.6pt}
\begin{tabular}{lccc}
\toprule
Method & $0.75\times$ & $1.25\times$ & Avg. warp\\
\midrule
NoFuture & $54.0\!\pm\!3.2$ & $54.0\!\pm\!3.2$ & $54.0$\\
GenFuture & $57.1\!\pm\!3.2$ & $60.1\!\pm\!4.1$ & $58.6$\\
Gen+RAFC & $\mathbf{70.9\!\pm\!2.6}$ & $\mathbf{75.4\!\pm\!1.9}$ & $\mathbf{73.2}$\\
\bottomrule
\end{tabular}
\end{table}

\subsection{H3: Natural Timing Mismatch on Hardware}
We evaluate opening a kettle, closing a kettle, and closing a microwave door
on a Franka under variable physical execution, where temporal mismatch may
arise naturally but is not measured. For each task, we collect 50--100
teleoperated demonstrations and train the future-conditioned BC policy.
We then train the residual actor and, for FEC+RAFC, the RAFC gate offline
using the same dataset. Both conditions use identical demonstrations,
offline training budgets, and generated futures; they differ in whether
RAFC is present. Plain FEC and FEC+RAFC are evaluated with matched
observations, initial conditions, horizons, and success criteria over
20 trials per task. Aggregate success rises from 16/60 (26.7\%) for
plain FEC to 34/60 (56.7\%) for FEC+RAFC
(Table~\ref{tab:real_robot}).

\begin{table}[!htbp]
\caption{Real-robot BC+RL evaluation over 20 trials per task.}
\label{tab:real_robot}
\centering
\scriptsize
\setlength{\tabcolsep}{4.0pt}
\begin{tabular}{lcc}
\toprule
Task & Plain FEC & FEC+RAFC \\
\midrule
Open kettle & 3/20 (15\%) & \textbf{9/20 (45\%)} \\
Close kettle & 5/20 (25\%) & \textbf{11/20 (55\%)} \\
Close microwave door & 8/20 (40\%) & \textbf{14/20 (70\%)} \\
\midrule
Aggregate & 16/60 (26.7\%) & \textbf{34/60 (56.7\%)} \\
\bottomrule
\end{tabular}
\end{table}

\subsection{Qualitative Future/Execution Consistency}
\begin{figure}[!t]
  \centering
  \includegraphics[width=\columnwidth]{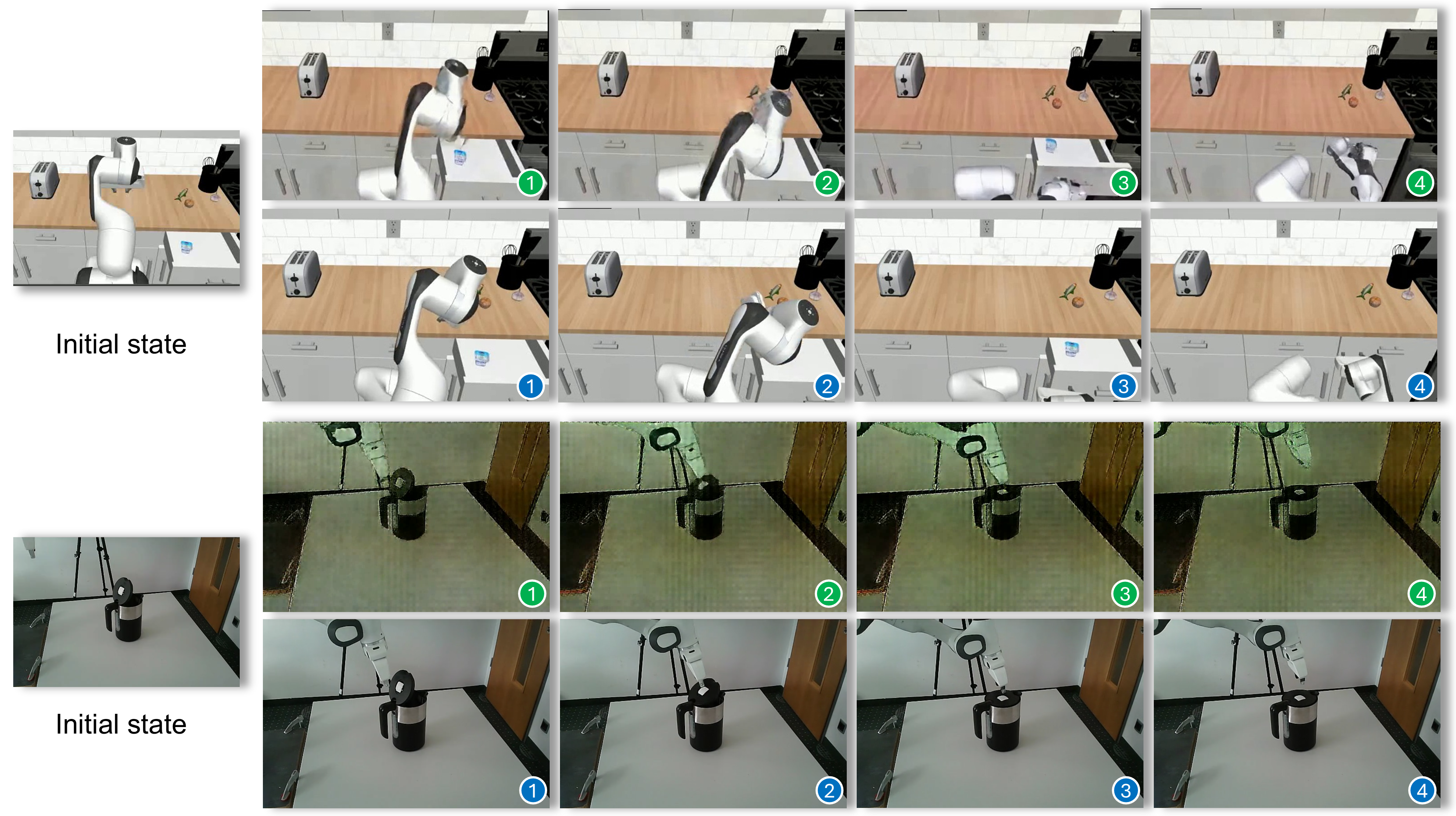}
  \caption{Generated futures (\textcolor{plangreen}{green}) and Policy (\textcolor{planblue}{blue}), read
  left to right. Top: RoboCasa drawer closing. Bottom: Franka
  kettle closing. Futures guide interaction, not exact trajectories.}
  \label{fig:qualitative}
\end{figure}

Figure~\ref{fig:qualitative} pairs generated conditioning futures with
closed-loop execution in simulation and hardware; futures specify the target
interaction rather than a reference trajectory.

\subsection{RL Fine-Tuning Efficiency}
\begin{figure}[!htbp]
  \centering
  \includegraphics[width=0.94\columnwidth]{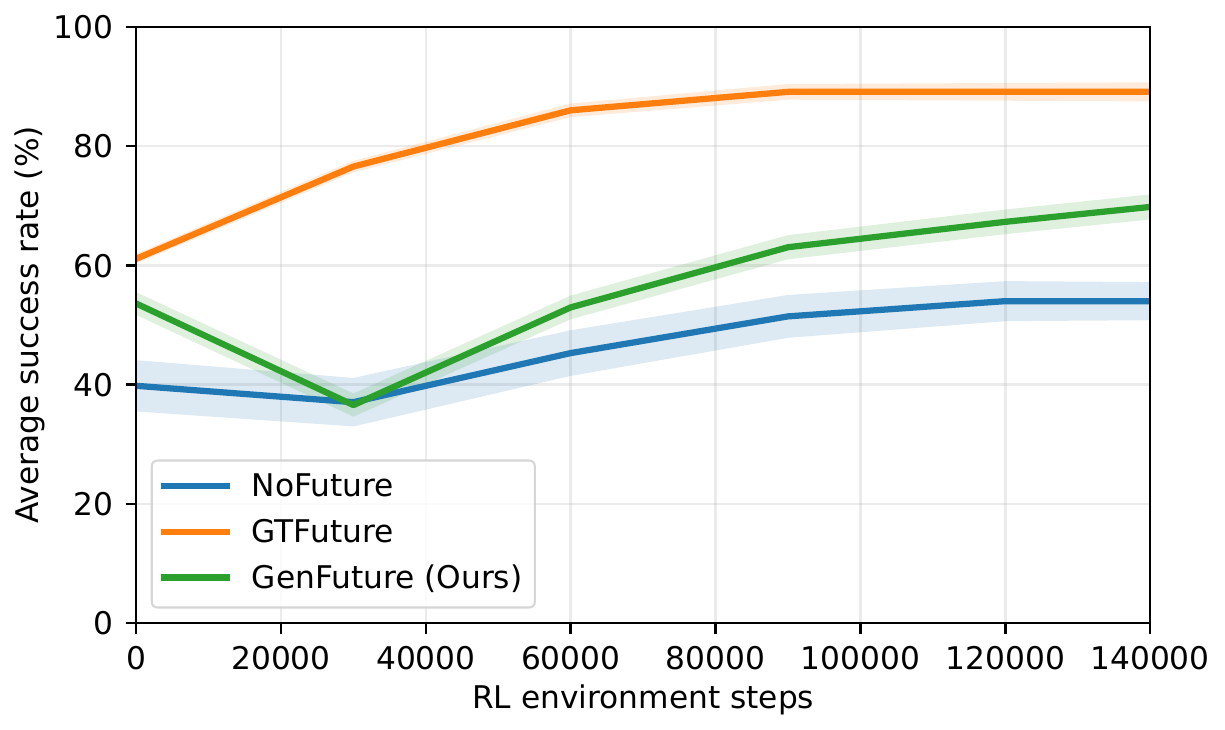}
  \caption{Average BC+RL learning curves across eight CALVIN tasks; step 0 is the corresponding BC-only policy.}
  \label{fig:learning}
\end{figure}

In Figure~\ref{fig:learning},
GTFuture reaches 70\% after approximately 17,260 steps;
GenFuture reaches 60\% after approximately 80,955 steps
and 69.8\% at 140k steps following an early dip, while
NoFuture reaches neither threshold within 140k steps.
Final reported success is 89.1\% with GTFuture,
69.8\% with GenFuture, and 54.0\% with NoFuture.

\section{Discussion}

\textbf{Learned reliability provides gain beyond ensembling.} On the off-grid
shifts, ShiftAug recovers 13.3 points over GenFuture, uniform averaging over
the identical candidate bank recovers 14.9, and full RAFC recovers 21.9.
Thus, candidate diversity alone provides a strong recovery, but learned
reliability contributes a further 7.0 points with the same candidate bank and
RL budget. This additional gain cannot be explained by exact candidate
matching, since every nonzero evaluation shift is deliberately off-grid.
The gate diagnostics and step-40 intervention show the corresponding mechanism:
weights move in the compensating direction, while large residual mismatch also
reduces $\alpha$.

\textbf{Generation and reliability are separable concerns.} The robot-free
twin specifies object articulation while video diffusion restores robot
appearance, separating object-state generation from embodiment-specific
motion. RAFC then operates entirely downstream of that pipeline, on frozen-BC
features, which is why it transfers unchanged to hardware, where success rises
from 16/60 (26.7\%) to 34/60 (56.7\%) under mismatch nobody imposed. A better
generator and a better reliability mechanism are complementary, and the
present results quantify the second while holding the first fixed.

\section{Limitations}
RAFC composes corrections from $\mathcal S=\{-2,0,+2\}$, so phase errors beyond
that span cannot be fully realigned; at the largest off-grid shifts $\pm5$ and
in the additional $\pm6$ stress test, the gate reduces trust and falls back
rather than fully recovering. The coefficient
$\alpha_t$ is a learned allocation, not a calibrated probability that the
future is correct nor an estimate of the applied shift. Controlled shifts and
rate warps are confined to simulation, and the rate test covers monotone speed
mismatch rather than nonstationary warps; the physical study contains natural
mismatch but does not measure its episode-specific magnitude, making it
evidence of improvement over plain FEC rather than of mechanism isolation. 
Distinct-frame windows remove boundary repetition from the
imposed global shift, while RAFC's internal local candidates may still repeat
boundary frames. The experiments establish substantial sensitivity to phase errors in both
directions but do not isolate direction-specific failure mechanisms. We fix the local candidate
bank to $\mathcal S=\{-2,0,+2\}$ and temporal pooling to $B=4$; sensitivity
to candidate-bank size and pooling granularity remains for future study. BC is
trained on demonstration futures; direct BC training on generated futures
remains untested. Finally,
task grounding assumes a calibrated task-relevant object model, and RAFC
addresses temporal reliability only, so it cannot repair an incorrect goal or
object, or a weak base policy.

\section{Conclusion}
A generated future that shows the right task at the wrong moment can be
substantially worse than no future at all. RAFC treats that as a control problem rather than a
generation problem. It learns, from task reward alone, how strongly to trust a
future, which nearby temporal hypothesis to prefer, and when to fall back
toward a static one. Off-grid shifts, independent distinct-frame windows, rate warps, ablations,
gate diagnostics, the online timing intervention, and a three-task Franka study
converge on the same conclusion. Future-conditioned control needs an explicit mechanism for temporal
reliability, and estimating that reliability before acting recovers most of
what timing mismatch takes away.
\section{Acknowledgment}
AI tools (ChatGPT) was used to assist in creating the small robotic arm graphic in figure~\ref{fig:rafc}, and \ref{fig:pipeline}, and drafting code scripts for the data visualization in Figure 4. All final outputs were reviewed and verified by the authors.

\balance
\footnotesize

\bibliographystyle{IEEEtran}
\bibliography{IEEEabrv,references}

@article{tezerjani2024rlslam,
  title={A survey on reinforcement learning applications in {SLAM}},
  author={Tezerjani, M. D. and Khoshnazar, M. and Tangestanizadeh, M. and Kiani, A. and Yang, Q.},
  journal={arXiv preprint arXiv:2408.14518},
  year={2024}
}

@inproceedings{finn2017foresight,
  title={Deep visual foresight for planning robot motion},
  author={Finn, C. and Levine, S.},
  booktitle={Proc. IEEE International Conference on Robotics and Automation (ICRA)},
  year={2017}
}

@article{ha2018world,
  title={World models},
  author={Ha, D. and Schmidhuber, J.},
  journal={arXiv preprint arXiv:1803.10122},
  year={2018}
}

@article{hafner2019planet,
  title={Learning latent dynamics for planning from pixels},
  author={Hafner, D. and Lillicrap, T. and Fischer, I. and Villegas, R. and Ha, D. and Lee, H. and Davidson, J.},
  journal={arXiv preprint arXiv:1811.04551},
  year={2019}
}

@article{hafner2020dreamer,
  title={Dream to control: Learning behaviors by latent imagination},
  author={Hafner, D. and Lillicrap, T. and Ba, J. and Norouzi, M.},
  journal={arXiv preprint arXiv:1912.01603},
  year={2019}
}

@article{khoshnazar2026enact,
  title={{LLM}-guided future hypotheses for horizon-aware exploration in multi-step robot manipulation},
  author={Khoshnazar, M. and Melnik, A. and Beetz, M.},
  journal={arXiv preprint arXiv:2605.29864},
  year={2026}
}

@inproceedings{luo2025grounding,
  title={Grounding video models to actions through goal-conditioned exploration},
  author={Luo, Y. and Du, Y.},
  booktitle={International Conference on Learning Representations (ICLR)},
  year={2025}
}

@inproceedings{bu2024closedloop,
  title={Closed-loop visuomotor control with generative expectation for robotic manipulation ({CLOVER})},
  author={Bu, Q. and Zeng, J. and Chen, L. and Yang, Y. and Zhou, G. and Yan, J. and Luo, P. and Cui, H. and Ma, Y. and Li, H.},
  booktitle={Advances in Neural Information Processing Systems},
  year={2024}
}

@article{black2023susie,
  title={Zero-shot robotic manipulation with pretrained image-editing diffusion models},
  author={Black, K. and Nakamoto, M. and Atreya, P. and Walke, H. and Finn, C. and Kumar, A. and Levine, S.},
  journal={arXiv preprint arXiv:2310.10639},
  year={2023}
}

@inproceedings{huey2025orca,
  title={Imitation learning from a single temporally misaligned video},
  author={Huey, W. and Wang, H. and Wu, A. and Artzi, Y. and Choudhury, S.},
  booktitle={International Conference on Machine Learning (ICML)},
  year={2025}
}

@inproceedings{ho2020ddpm,
  title={Denoising diffusion probabilistic models},
  author={Ho, J. and Jain, A. and Abbeel, P.},
  booktitle={Advances in Neural Information Processing Systems},
  year={2020}
}

@inproceedings{ho2022video,
  title={Video diffusion models},
  author={Ho, J. and Salimans, T. and Gritsenko, A. and Chan, W. and Norouzi, M. and Fleet, D. J.},
  booktitle={Advances in Neural Information Processing Systems},
  year={2022}
}

@inproceedings{lugmayr2022repaint,
  title={{RePaint}: Inpainting using denoising diffusion probabilistic models},
  author={Lugmayr, A. and Danelljan, M. and Romero, A. and Yu, F. and Timofte, R. and Van Gool, L.},
  booktitle={Proc. IEEE/CVF Conference on Computer Vision and Pattern Recognition (CVPR)},
  year={2022}
}

@article{yang2024cogvideox,
  title={{CogVideoX}: Text-to-video diffusion models with an expert transformer},
  author={Yang, Z. and others},
  journal={arXiv preprint arXiv:2408.06072},
  year={2024}
}

@article{bian2025videopainter,
  title={{VideoPainter}: Any-length video inpainting and editing with plug-and-play context control},
  author={Bian, Y. and Zhang, Z. and Ju, X. and Cao, M. and Xie, L. and Shan, Y. and Xu, Q.},
  journal={arXiv preprint arXiv:2503.05639},
  year={2025}
}

@inproceedings{kapelyukh2024dream2real,
  title={{Dream2Real}: Zero-shot 3D object rearrangement with vision-language models},
  author={Kapelyukh, I. and Ren, Y. and Alzugaray, I. and Johns, E.},
  booktitle={Proc. IEEE International Conference on Robotics and Automation (ICRA)},
  pages={4796--4803},
  year={2024}
}

@article{ahn2022saycan,
  title={Do as I can, not as I say: Grounding language in robotic affordances},
  author={Ahn, M. and others},
  journal={arXiv preprint arXiv:2204.01691},
  year={2022}
}

@inproceedings{mandlekar2022robomimic,
  title={What matters in learning from offline human demonstrations for robot manipulation},
  author={Mandlekar, A. and others},
  booktitle={Conference on Robot Learning (CoRL)},
  year={2022}
}

@inproceedings{fujimoto2018td3,
  title={Addressing function approximation error in actor-critic methods},
  author={Fujimoto, S. and Hoof, H. and Meger, D.},
  booktitle={International Conference on Machine Learning (ICML)},
  year={2018}
}

@article{nair2020awac,
  title={{AWAC}: Accelerating online reinforcement learning with offline datasets},
  author={Nair, A. and Gupta, A. and Dalal, M. and Levine, S.},
  journal={arXiv preprint arXiv:2006.09359},
  year={2020}
}

@inproceedings{lee2022offlineonline,
  title={Offline-to-online reinforcement learning via balanced replay and pessimistic {Q}-ensemble},
  author={Lee, S. and Seo, Y. and Lee, K. and Abbeel, P. and Shin, J.},
  booktitle={Conference on Robot Learning (CoRL)},
  year={2022}
}

@article{jiang2025sfp,
  title={Streaming flow policy: Simplifying diffusion/flow-matching policies by treating action trajectories as flow trajectories},
  author={Jiang, S. and Fang, X. and Roy, N. and Lozano-Perez, T. and Kaelbling, L. P. and Ancha, S.},
  journal={arXiv preprint arXiv:2505.21851},
  year={2025}
}

@inproceedings{chi2023diffusion,
  title={Diffusion policy: Visuomotor policy learning via action diffusion},
  author={Chi, C. and Xu, Z. and Feng, S. and Cousineau, E. and Du, Y. and Burchfiel, B. and Tedrake, R. and Song, S.},
  booktitle={Robotics: Science and Systems (RSS)},
  year={2023}
}

@article{mees2022calvin,
  title={{CALVIN}: A benchmark for language-conditioned policy learning for long-horizon robot manipulation tasks},
  author={Mees, O. and Hermann, L. and Rosete-Beas, E. and Burgard, W.},
  journal={IEEE Robotics and Automation Letters},
  volume={7},
  number={3},
  pages={7327--7334},
  year={2022}
}

@inproceedings{nasiriany2024robocasa,
  title={{RoboCasa}: Large-scale simulation of everyday tasks for generalist robots},
  author={Nasiriany, S. and Maddukuri, A. and Zhang, L. and Parikh, A. and Lo, A. and Joshi, A. and Mandlekar, A. and Zhu, Y.},
  booktitle={Robotics: Science and Systems (RSS)},
  year={2024}
}

@inproceedings{zhou2023propainter,
  title={{ProPainter}: Improving propagation and transformer for video inpainting},
  author={Zhou, S. and Li, C. and Chan, K. C. K. and Loy, C. C.},
  booktitle={Proc. IEEE/CVF International Conference on Computer Vision (ICCV)},
  pages={10477--10486},
  year={2023}
}

@inproceedings{du2023unipi,
  title={Learning universal policies via text-guided video generation},
  author={Du, Yilun and Yang, Sherry and Dai, Bo and Dai, Hanjun and Nachum, Ofir and Tenenbaum, Joshua B. and Schuurmans, Dale and Abbeel, Pieter},
  booktitle={Advances in Neural Information Processing Systems},
  volume={36},
  year={2023}
}

@inproceedings{ko2024avdc,
  title={Learning to act from actionless videos through dense correspondences},
  author={Ko, Po-Chen and Mao, Jiayuan and Du, Yilun and Sun, Shao-Hua and Tenenbaum, Joshua B.},
  booktitle={International Conference on Learning Representations},
  year={2024}
}

@inproceedings{dwibedi2019tcc,
  title={Temporal cycle-consistency learning},
  author={Dwibedi, Debidatta and Aytar, Yusuf and Tompson, Jonathan and Sermanet, Pierre and Zisserman, Andrew},
  booktitle={Proc. IEEE/CVF Conference on Computer Vision and Pattern Recognition (CVPR)},
  pages={1801--1810},
  year={2019}
}

@inproceedings{zakka2022xirl,
  title={{XIRL}: Cross-embodiment inverse reinforcement learning},
  author={Zakka, Kevin and Zeng, Andy and Florence, Pete and Tompson, Jonathan and Bohg, Jeannette and Dwibedi, Debidatta},
  booktitle={Conference on Robot Learning (CoRL)},
  year={2022}
}

@inproceedings{ma2023vip,
  title={{VIP}: Towards universal visual reward and representation via value-implicit pre-training},
  author={Ma, Yecheng Jason and Sodhani, Shagun and Jayaraman, Dinesh and Bastani, Osbert and Kumar, Vikash and Zhang, Amy},
  booktitle={International Conference on Learning Representations (ICLR)},
  year={2023}
}

@inproceedings{johannink2019residual,
  title={Residual reinforcement learning for robot control},
  author={Johannink, Tobias and Bahl, Shikhar and Nair, Ashvin and Luo, Jianlan and Kumar, Avinash and Loskyll, Matthias and Ojea, Juan Aparicio and Solowjow, Eugen and Levine, Sergey},
  booktitle={Proc. IEEE International Conference on Robotics and Automation (ICRA)},
  pages={6023--6029},
  year={2019}
}

@article{ankile2024resip,
  title={From imitation to refinement: Residual {RL} for precise assembly},
  author={Ankile, Lars and Simeonov, Anthony and Shenfeld, Idan and Torne, Marcel and Agrawal, Pulkit},
  journal={arXiv preprint arXiv:2407.16677},
  year={2024}
}

\end{document}